\documentclass[letterpaper, 10 pt, conference]{ieeeconf}  

\usepackage{xcolor}
\usepackage{amssymb}
\usepackage{amsmath}
\usepackage{graphicx}
\usepackage{algorithm}
\usepackage{algorithmic}
\usepackage{indentfirst}
\usepackage{dsfont}
\usepackage{subfigure}
\usepackage{booktabs}
 \usepackage{multirow}

\IEEEoverridecommandlockouts                              

\usepackage{eso-pic}
\usepackage{xcolor}

\AddToShipoutPictureBG*{
  \AtPageUpperLeft{
    \put(\LenToUnit{0.5\paperwidth},\LenToUnit{-1.5cm}){ 
      \makebox[0pt][c]{
        \color{gray} 
        \small 
        \begin{tabular}{c}
          This work has been accepted for publication at the \\
          IEEE International Conference on Robotics and Automation (ICRA), 2026, \copyright IEEE
        \end{tabular}
      }
    }
  }
}
\title{\LARGE \bf
Learning to Annotate Delayed and False AEB Events: 
A Practical System for Extreme Class Imbalance and Asymmetric Label Noise 
}
\author{Mengxiang Hao$^{*}$, Xin Jiang, Xinghao Huang, Wenliang Su, Zhiteng Wang, Junjie Rao, \\
Xiaotian Yang, Wei Liao, Chengyu Han, Gen Liang, Yulun Song, Zhitao Xu, Xianpeng Lang
\thanks{All authors are with Li Auto, China.}%
\thanks{$^{*}$Corresponding author. E-mail: {\tt\small haomengxiang@lixiang.com}}%
}

\begin{document}

\maketitle
\thispagestyle{empty}
\pagestyle{empty}

\begin{abstract}
Autonomous Emergency Braking (AEB) optimization relies on accurately annotated real-world trigger events, particularly rare but critical delayed and false AEB triggers that expose system deficiencies. However, these minority samples comprise less than 5\% of thousands of daily triggers, making manual annotation prohibitively expensive at scale. We present the first automated AEB annotation framework to address this problem. During development, we identified two fundamental challenges that severely impair delayed/false trigger annotation accuracy: (1) Extreme class imbalance where delayed/false triggers are overwhelmed by true triggers; (2) Asymmetric label noise where mislabeled majority samples (true triggers) suppress minority samples (delayed/false triggers) learning. To overcome these challenges, we propose two key innovations: (1) Specific data augmentation that synthesizes realistic samples by manipulating focal target attributes, transplanting ego-vehicle dynamics, and masking non-focal agents; (2) noise suppression using stable hardness estimation and probe-guided adaptive threshold to clean mislabeled true trigger samples. Crucially, we deploy our model as a practical annotation system with full-stack architecture, efficiently identifying critical delayed/false triggers from thousands of daily AEB events. Production results demonstrate 80\% improvement in recall of delayed/false triggers and 50\% reduction in manual workload. Beyond immediate gains, the system enables continuous self-improvement through accumulated high-quality annotations, establishing a necessary data foundation for on-vehicle AEB system optimization.
\end{abstract}


\section{INTRODUCTION}
Autonomous Emergency Braking (AEB) is a  driver-assist system that senses an imminent collision and, without driver input, applies the brakes to cut speed and, if possible, prevent the impact \cite{EuroNCAP_AEB_VRU_2024}. A 2015 study found that cars equipped with AEB suffered 38\% fewer rear-end crashes, underscoring the system’s effectiveness \cite{fildes2015effectiveness}.

Optimizing AEB hinges on data collected from real-world activations—without detected and labeled delayed/false AEB trigger events, system improvement becomes impossible. The AEB events are simply tagged as: 
True trigger - on time, collision avoided;
False trigger - brakes with no threat;
Delayed trigger - brakes engage late, leaving too little stopping distance and a high collision risk. Although infrequent, false and delayed triggers are crucial because their labels pinpoint where the perception or control logic needs refinement—delayed triggers expose missed threats, while false triggers reveal over-reactive detection or algorithm defects. Consequently, the primary objective of annotation is to accurately identify false and delayed triggers within vast amounts of driving data, thereby providing engineers with precise insights to optimize system performance under the corresponding conditions.

However, after years of system optimization, both delayed and false triggers now comprise less than 5\% of the thousands of daily trigger events and AEB-equipped vehicle populations grow exponentially, making manual analysis of this massive data stream expensive. This motivates the development of automated annotation systems that can efficiently process large scale data while maintaining high accuracy for safety-critical delayed/false trigger events.

\textbf{We develop the first AEB annotation model.} During model development, we encountered two fundamental challenges that critically impair the annotation of delayed/false triggers in real-world AEB data:

Challenge 1 - Class imbalance: The delayed/false triggers are exceedingly scarce relative to true triggers, 
which causes models to be overwhelmingly biased toward predicting the majority class (true triggers), as the loss contribution from minority class (delayed/false triggers) becomes negligible during training \cite{johnson2019survey}. Consequently, the model fails to learn discriminative features for delayed/false triggers.

Challenge 2 - Asymmetric label noise: Real AEB annotation exhibits a unique asymmetric pattern where delayed/false triggers (requiring expert careful 
analysis) are near-perfectly labeled, while the true triggers 
contain around 1\% mislabeled samples (samples labeled as true triggers but actually delayed/false triggers). 
When the model encounters minority features but sees them labeled as majority, it learns to suppress minority class predictions to minimize training loss \cite{natarajan2013learning}. As a result, the model becomes increasingly conservative, preferring the "safe" majority class prediction even when facing genuine minority instances. 

To tackle the aforementioned problems, we survey existing methods in the fields of imbalanced classification and long-tailed classification \cite{tan2020equalization,liu2025climb}. While these existing learning methods have shown success in various domains, they cannot be directly applied to AEB annotation due to the unique characteristics of AEB data \cite{chawla2002smote,zhang2022self}. Although some general methods \cite{lin2017focal, ridnik2021asymmetric,lakshminarayanan2017simple} are applicable to AEB data, they yield suboptimal results due to the unique combination of extreme imbalance and asymmetric noise.

Consequently, we incorporate \textbf{two key technical innovations} to address these challenges:
First, we propose \textbf{AEB-targeted data augmentation strategies} for delayed/false triggers to address the extreme class imbalance, which synthesize realistic delayed/false trigger samples by manipulating focal target attributes, transplanting ego-vehicle vectors and masking non-focal agents.
Second, we designed a \textbf{noise suppression mechanism}, which use stable hardness estimation across training epochs and probe-guided adaptive thresholds to distinguish mislabeled samples from borderline cases in majority class, enabling more confident detection of delayed/false triggers.

\textbf{Beyond individual model, we build a production system around it and deliver a complete annotation pipeline.} 
Our system implements a full-stack architecture, enabling seamless human‑AI collaboration that bridges the gap between theoretical models and a practical annotation system capable of meeting stringent AEB requirements at scale.

Our main contributions are as follows. 
\begin{itemize}
    \item \textbf{First AEB annotation model:} We develop the annotation model specifically designed for AEB trigger classification, addressing the unique temporal-spatial characteristics of AEB. 
    \item \textbf{Novel dual-strategy framework for real data challenges:} (1) AEB-targeted augmentation that leverages domain physics to synthesize realistic delayed/false triggers, addressing extreme class imbalance; (2) Noise suppression that guided by probe noise samples' hardness to adaptively identifies mislabeled majority samples, mitigating asymmetric label noise. Experimental validation demonstrates that each method contributes to enhanced delayed/false triggers learning.
    \item \textbf{Deployed annotation system at production scale:} We develop a full-fledged annotation system in which our proposed model serves as the core, which combines automated processing with human verification for thousands of daily events. Production deployment demonstrates 80\% improvement in minority class detection and 50\% workload reduction. Beyond immediate gains, the system leverages its continuously accumulated annotation data—both human-verified minority cases and automatically cleaned majority samples—to periodically retrain and improve the model, establishing a self-evolving framework.
\end{itemize}

\section{Related Work}

As AEB data annotation is largely an industrial task with no academic studies to the best of our knowledge, our literature review concentrates on the two data challenges noted above: (1) classification under severe class imbalance and (2) classification in the presence of label noise.
\subsection{Imbalance Classification Models Problem}
Researches \cite{de2024survey,zhang2023deep} have demonstrated that the imbalance of training samples leads to poor performance on tail data. However, AEB annotation models are inherently supposed to focus on this part of tail data related to false activations and collision; Thus, the imbalance problem is an inevitable challenge that must be addressed.

Remedies for class imbalance are typically pursued from three angles \cite{zhang2023deep, liu2025climb}: data-level manipulation, algorithmic modification, and ensemble techniques.
\begin{enumerate}

\item At the data level, the principal method is resampling~\cite{chawla2002smote, estabrooks2004multiple, liu2008exploratory, zhang2021learning}, which comes in two directions: oversampling and undersampling. Oversampling duplicates instances from the minority class, whereas undersampling discards a portion of the majority class data, so that the class proportions in the training set are kept within a reasonable range. The effectiveness of resampling has been empirically demonstrated \cite{estabrooks2004multiple, liu2008exploratory}. Chawla et al.\cite{chawla2002smote} introduced the Synthetic Minority Over-sampling Technique (SMOTE), which generates additional minority samples by interpolating between existing ones, thus improving classifier performance on imbalanced datasets. For undersampling, Edited Nearest Neighbors (ENN)~\cite{wilson1972asymptotic} removes majority class samples that are misclassified by their k-nearest neighbors, effectively cleaning the decision boundary while reducing class imbalance.

\item At the algorithm level, imbalance bias is usually alleviated by modifying the loss function or its class-specific weights~\cite{ridnik2021asymmetric,lin2017focal, cui2019class, ren2020balanced}. A representative example is focal loss \cite{lin2017focal}, which shifts the learning focus toward hard, low-frequency classes.  Besides, Ridnik et al.\cite{ridnik2021asymmetric} propose Asymmetric Loss that decouples the contributions of positive and negative samples, applying asymmetric focusing to handle the inherent imbalance in multi-label classification tasks.
 
\item Ensemble methods address the challenge by training a set of specialized expert models, each tailored to a particular region or data scenario, and then combining their output to achieve a strong overall classification performance \cite{zhou2020bbn, liu2020self, zhang2022self}. For example, Deep Ensemble~\cite{lakshminarayanan2017simple} trains multiple networks with different initializations and averages their predictions. 
The SADE~\cite{zhang2022self} trains diverse experts on different class distributions and aggregates them using self-supervision without knowing test distribution. 
\end{enumerate}

However, the reweight and ensemble methods are general but has limited effect; The existing resample methods are not suitable for AEB task; Besides, these methods primarily optimize balanced metrics that weight all classes equally, whereas AEB annotation places greater emphasis on maximizing the performance on delayed/false triggers. Consequently, a bespoke solution has to be developed.

\subsection{Noise Label Detection}
Naive rebalance is not enough, because data quality impacts results more severely than imbalance itself\cite{kang2019decoupling,liu2025climb}. Northcutt et al.\cite{northcutt2021pervasive} highlight that label noise can markedly degrade network accuracy, making noise mitigation essential for reliable models.

O2U-Net\cite{huang2019o2u} addresses label noise by leveraging prediction uncertainty during training. It identifies potentially mislabeled samples based on their prediction inconsistency across epochs—samples with high uncertainty are likely noisy. This simple yet effective approach requires no additional model complexity. Industrial applications like Masking-Mix\cite{nishi2021augmentation} and Prototypical-Cleanser\cite{wei2022prototypical} have adopted similar uncertainty-based strategies to improve training data quality.

In conclusion, existing methods address either imbalance or noise in isolation, but AEB annotation faces both extreme class imbalance and asymmetric label noise. This unique challenge requires a dual strategy.

\section{Methodology}
\label{sec:method}

This section introduces a methodology for tackling data imbalance and asymmetric label noise as discussed above. It starts with task modeling, proceeds to two complementary strategies, and concludes with a sequential training workflow uniting all components.

\subsection{Task Modeling for AEB Data Annotation}
\label{sec:model}
Since AEB events occur in one of three forms—\textbf{true}, \textbf{false}, or \textbf{delayed}, we cast the annotation problem as a three class classification task.

\subsubsection{Data Source and Feature Construction}
The activation of AEB is determined by the dynamic state information of surrounding agents and the ego-vehicle. We extract the agent and ego-vehicle feature inputs from real-vehicle data streams, which are first downsampled to 20 Hz for temporal alignment and then centered on the AEB trigger moment. This preprocessing strategy captures the spatiotemporal interaction between surrounding agents and the ego-vehicle before and after AEB activation, which is critical for judging the necessity and timeliness of the trigger event.

\paragraph{Agent-Level Feature} 
The feature comes from upstream perception modules, matching the format of 3D perception tracking boxes in the on-vehicle AEB system. Each box includes \textit{Spatial Features}, the relative center coordinates $(x, y, z)$ (w.r.t. the ego-vehicle), dimensions $(l, w, h)$,  and yaw angle $\theta$ as well as \textit{Kinematic Features}, the velocity $(v_x, v_y)$, acceleration $(a_x, a_y)$, and yaw rate $\omega$.  

To standardize input dimensions, the number of obstacles is capped at \textit{A=32}: if more than \textit{A} exist, the \textit{A} closest to the ego-vehicle are retained; if fewer, zero-padding is applied. 

\paragraph{Ego-Vehicle Feature}  
Ego-vehicle features are from on-vehicle sensors include spatial and kinematic attributes as the same as agents, additional with \textit{control signals}: throttle position, brake pressure, steering angle and the \textbf{AEB activation flag}, a binary indicator marking trigger start and end moments.

\subsubsection{Transformer-based Model Architecture for Temporal-Spatial Encoding}
\begin{figure}[t]  
\centering
\includegraphics[width=\linewidth, trim=1pt 1pt 1pt 1pt, clip]{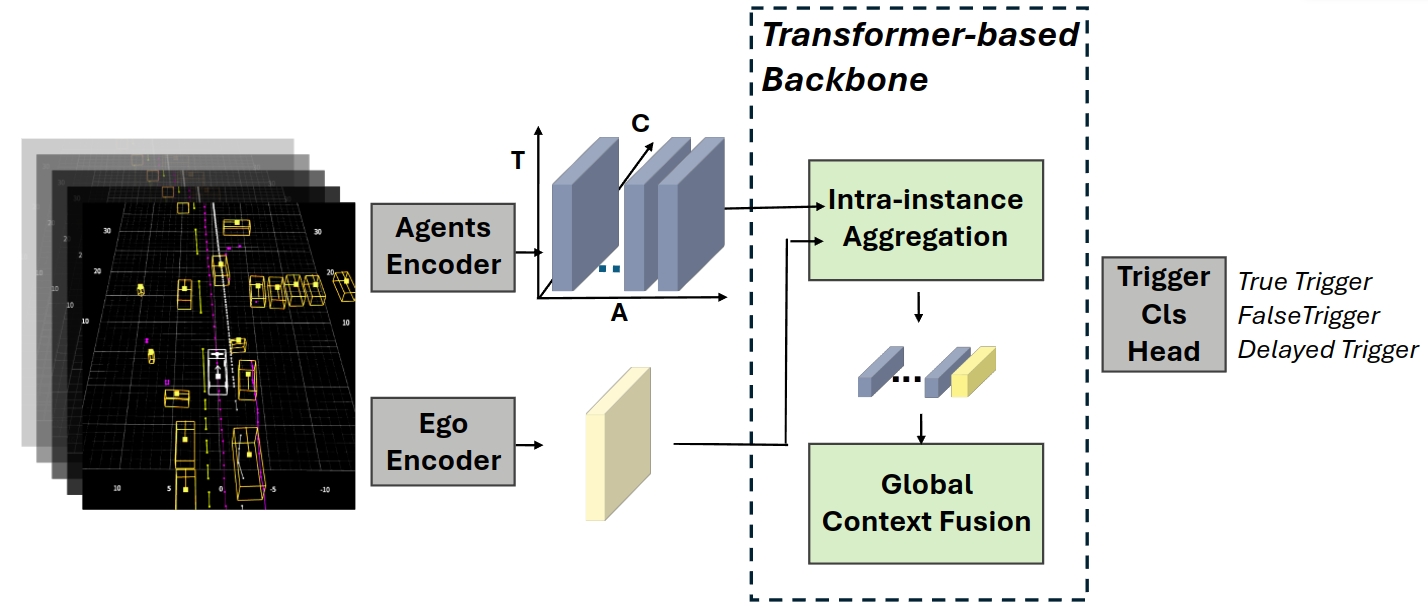}  
\caption{The Architecture of our AEB annotation model.}
\label{fig:model}
\end{figure}

To model the spatiotemporal context of AEB events, our model adopts a unified embedding module and a Transformer backbone architecture as shown in Figure~\ref{fig:model}.

\paragraph{Unified Embedding Module}  
This module aligns ego and agent features into a shared space while preserving critical information. Both agent and ego features are processed through their dedicated encoders, which internally incorporate sinusoidal timestamp encoding modules to encode relative time information. After encoding, both streams are mapped to dimension C,  forming tensors \(A \times T \times C\) and \(1 \times T \times C\).

\paragraph{Transformer Backbone and Classification Head} 
We use the Transformer to first conduct intra-instance feature aggregation, which performs attention operations along the time dimension, then compresses this dimension, yielding a tensor of shape \((1 + A) \times C\). Then, attention is applied across the \(1 + A\) dimension, followed by maxpooling to extract global context features.
Finally, the processed features are passed to the \textit{Trigger Cls Head} to output classification results.

Building on the above process, we establish a foundational model for AEB annotation.
However, due to the poor distribution, the direct application of the aforementioned model yields unsatisfactory minority performance. Subsequently, we propose the following AEB-targeted augmentation method to alleviate the imbalance problem and a noise suppression mechanism to adaptively separate noise from  true triggers.

\subsection{ Targeted Augmentation Method for Delayed/False Triggers}
\label{subsec:oversample}
For the first challenge, we choose oversampling as our direction because undersampling would discard most of true triggers on our extreme imbalance condition causing unacceptable information loss, while reweighting merely shifts model focus without addressing the lack of minority class.

\begin{figure}[t]  
\centering
\includegraphics[width=\linewidth, trim=10pt 10pt 10pt 10pt, clip]{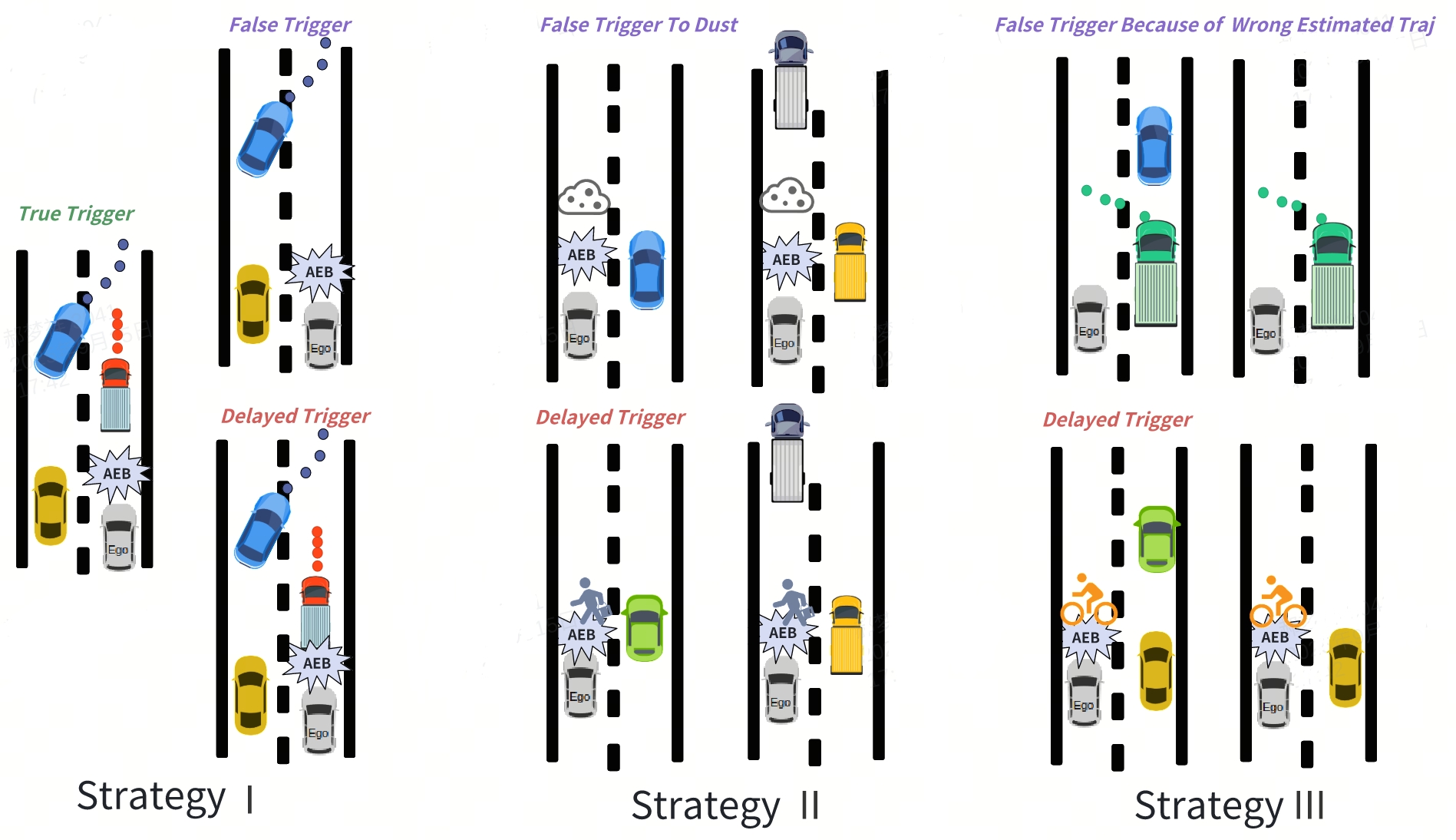}  
\caption{Generated delayed/false trigger events by our strategies. \textbf{Strategy I:} top: removing the focal target via zero-masking focal feature vector; bottom: shortening the focal target’s longitudinal distance by adjusting its position to mimic delayed triggers. \textbf{Strategy II:} transplanting ego-vehicle and focal target attribute vectors from real minority samples into true trigger scenarios. \textbf{Strategy III:} randomly masking non-focal obstacle vectors to add background variability while preserving core trigger logic.}
\label{fig:data-aug}
\end{figure}

The production rule-based AEB system first selects the \textbf{focal target} and then calculates the Time To Collision (TTC) and other conditions based on focal target's relative motion trend of the vehicle to determine whether to trigger AEB. The attributes of focal target directly determine the necessity and timeliness of trigger events. Leveraging this core characteristic, we propose three AEB-targeted data augmentation strategies to generate more delayed/false samples. The generated samples are shown as Figure~\ref{fig:data-aug}.

\subsubsection{\textbf{Strategy \uppercase\expandafter{\romannumeral1}, Change Focal Target Attribute}}
\label{subsubsec:s1}
We modify key attributes of the focal target of true trigger samples to simulate delayed and false trigger scenarios.

\textit{Delayed Trigger}: To mimic insufficient braking distance due to delayed AEB activation, we retrieve the longitudinal distance \(d_{end}\) between the ego-vehicle and the focal target when the AEB trigger ends. Subsequently, we adjust the longitudinal position in the focal target’s feature vector by randomly subtracting \(d_{end} \times (50\% - 150\%)\). Consequently, in the synthesized data, the ego-vehicle either stops at an abnormally short distance from the target or even collides with it, consistent with the core risk of delayed triggers. 

\textit{False Trigger}: We propose two methods to convert true trigger to false trigger. First, the lateral position attribute of the focal target is shifted away by a distance equivalent to one vehicle width. After AEB braking, the target no longer lies within the ego-vehicle’s potential collision zone, becoming an unnecessary false trigger. Secondly, The entire focal feature vector is masked as zero. This directly simulates scenarios where the AEB system triggers braking in response to a non-existent target.

\subsubsection{\textbf{Strategy \uppercase\expandafter{\romannumeral2}, Ego-Focal Vector Transplantation}}

Real-world false and delayed triggers often stem from complex error scenarios, such as dust false detection and undetected focal target due to occlusion which are difficult to fully simulate as Strategy \uppercase\expandafter{\romannumeral1}.

To capture these real-world characteristics, we design an ego-focal vector transplantation strategy: We first extract and store the ego-vehicle and focal target attribute vectors from actual delayed/false trigger samples, forming a dedicated attribute bank. We then replace both the ego and focal target vectors in true trigger samples with randomly selected vector pairs from this bank. This operation is facilitated by our input data format, an \(A \times T \times C\) tensor where the midpoint of \(T\) corresponds to the AEB trigger moment, allowing direct vector replacement between different samples without additional time alignment. By transplanting real error-related attribute patterns, this strategy generates more authentic delayed/false trigger samples that reflect real-world failure modes.

\subsubsection{\textbf{Strategy \uppercase\expandafter{\romannumeral3}, Non-Focal Target Random Masking}}
We randomly mask the non-focal obstacles vector of real minority class data. This perturbation introduces reasonable variability in background traffic scenarios but does not affect the core trigger logic, ensuring the augmented samples remain representative of real AEB minority class cases.  

\subsubsection{\textbf{Plausibility Validation}}
After generating synthetic samples via Strategies \uppercase\expandafter{\romannumeral1}/\uppercase\expandafter{\romannumeral2}, we calculate the IoU between focal and other agents in the scene. If any intersection is detected, indicating a violation of real traffic physics. Only samples that pass this validation are retained as valid generated training data.  

\subsubsection{\textbf{Constraints of Generated AEB Data}}
While the aforementioned strategies efficiently generate large volumes of delayed/false triggers, two key limitations persist in the generated data, affecting its authenticity and coverage of real-world scenarios.

First, synthetic sample authenticity is limited. Our plausibility check only validates spatial constraints, missing the complex dynamic interactions between agents that would naturally change when focal target attributes are modified.

Second, the strategies fail to simulate some typical real-world delayed/false trigger scenarios. 
For instance, 
delayed triggers induced by \textbf{focal target track ID jumps}. In practice, the AEB system requires to verify the agent's existence time and the perception system sometimes exhibits unstable detection performance. If the target’s track ID jumps, which is unavoidable even with the stable Kalman filter-based tracking we adopt, the verification process restarts, leading to delayed triggering. However, track ID jumps cause the original focal target to be split into multiple independent agent vectors in the input agents tensor, making it difficult to simulate.

To avoid overfitting the model to artificially scenarios, we strictly control the ratio of synthetic minority samples to real minority samples at approximately $1:1$. 

\subsection{Hardness-driven Noise Suppression for Majority Class}
\label{subsec:noise_suppression}

For the second challenge, our goal is to identify and suppress noisy samples (actually delayed/false triggers but mislabeled as true triggers) from the majority class while preserving valuable borderline true trigger samples.

\subsubsection{Hardness-based Sample Characterization}

The key insight is that mislabeled samples exhibit distinct training behaviors compared to correctly labeled samples. When a delayed/false trigger is mislabeled as true trigger, the model struggles to learn this contradictory pattern throughout training, whereas genuine borderline samples may be initially difficult but gradually become learnable.

To capture this distinction, we monitor each sample's prediction confidence during training. For a sample with incorrect label, the model's prediction will persistently deviate from the given label, resulting in consistently low confidence. We quantify this behavior through a \textbf{hardness} metric:


\begin{equation}
H_i = |1 - p_i|
\end{equation}
where $p_i$ is the model's confidence for the $i$-th sample's true trigger prediction. This metric effectively captures how "hard" it is for the model to fit each sample to its given label, while mislabeled samples maintain high hardness values as the model resists learning incorrect patterns.


Based on our quantitative and qualitative analysis in Section~\ref{subsubsec:hardness_exp}, we observe that hardness distributions can be unstable. To obtain robust hardness estimates, we apply exponential moving average (EMA) to stabilize hardness:

\begin{equation}
H_i^{(e)} = (1 - \alpha) \cdot H_i^{(e-1)} + \alpha \cdot |1 - p_i^{(e)}|
\end{equation}
where $e$ means the epoch, and we set $\alpha$ a small value (0.05) to preserve early-epoch signals when the distinction between noise and borderline samples is most pronounced.

\subsubsection{Probe-Guided Adaptive Threshold Selection}

While hardness effectively characterizes sample difficulty, determining an appropriate threshold to separate noise from borderline samples remains challenging. Setting the threshold too low risks removing valuable borderline samples; too high allows noise to persist. To address this, we propose a probe-guided approach that automatically calibrates the threshold based on known noise patterns.

\textbf{Probe Set Construction:} We construct \textit{Noise probes} by deliberately mislabeling minority samples to true triggers. These probes serve as reference points to understand how actual noise behaves during training.

\textbf{Adaptive Threshold Determination:} During training, we monitor the hardness distributions of probe data to learn the characteristic hardness range of noisy samples. The threshold for noise identification is set as:
\begin{equation}
\tau = \text{mean}(H_{\text{noise}}) + \epsilon
\end{equation}
where $\text{mean}(H_{\text{noise}})$ and $\epsilon$ are the mean and standard deviation of noise probe samples. This data-driven approach ensures the threshold adapts to the specific noise patterns in our AEB data.

Samples with $H_i > \tau$ are identified as likely mislabeled and subjected to discard. As demonstrated in our experimental analysis (Section~\ref{subsubsec:hardness_exp}), this probe-guided mechanism effectively distinguishes noise from borderline samples, achieving robust noise suppression without sacrificing valuable training data.

\subsection{Sequential Training Workflow}  
\begin{enumerate}
    \item Train the first base model using the full dataset, and acquire the hardness of all samples based on this training process. 
    \item For subsequent models (2nd to \( K \)-th):  
    Based on noise samples' hardness distribution to discard likely mislabeled true triggers. Retain all real minority class samples, and additionally sample synthetic minority class samples in equal proportion to the real ones. The 2nd to \( K \)-th  models are trained on distinct data samples, combined with random seed initializations and independent training processes \cite{lakshminarayanan2017simple}.
    \item Ensemble the 2nd to \( K \)-th models, and set a reasonable confidence threshold based on the ensemble models' performance on the validation set to ensure the recall of minority class samples. 
\end{enumerate}

\section{Experiments}
\label{sec:experiments}
\subsection{Experimental Setup}
\label{subsec:setup}

\textbf{Dataset:}
Experiments are conducted on a large scale AEB trigger dataset which contains around 100,000 manually annotated samples. Our dataset comprehensively covers the vast majority of real-world AEB trigger events for vulnerable road user (VRU) such as pedestrians, cyclists, and motorcyclists, as well as cars, etc., covering diverse traffic conditions. The ratio of true triggers to false triggers and delayed triggers is approximately $50:1:1$. Among all labeled data, 80\% is divided into the training set, 10\% into the validation set, and the remaining 10\% into the test set. 

\textbf{Training and Inference Details:}
For training, we sequentially train $K=4$ models and only adopt the last three models for ensemble. Training parameters are set as follows: Adam optimizer with an initial learning rate of $10^{-4}$, cosine annealing learning rate scheduler, batch size of 256, and 32 training epochs per base model. Training and inference relies on 4 NVIDIA A10 GPUs. 

\subsection{Evaluation Metric}
Based on AEB minority-recall-first principle, we select Delayed Trigger Precision@Recall=90\% (abbreviated as DP@R90\%) and False Trigger Precision@Recall=90\% (abbreviated as FP@R90\%) as our evaluation metric, which mean the precision when model recall 90\% corresponding class. The corresponding confidence threshold is set based on the model’s validation set performance. For trigger event that the confidence exceeds both delayed and false trigger threshold, the label with the higher confidence is selected. All experimental results presented below are evaluated on the test set.


\begin{figure*}[t]  
\centering
\includegraphics[width=\linewidth, trim=1pt 1pt 1pt 1pt, clip]{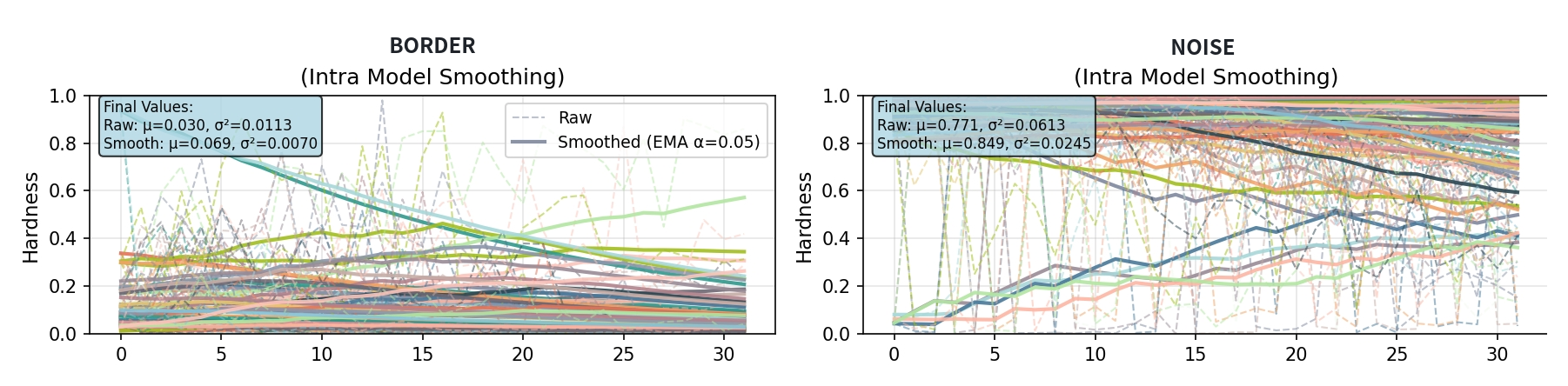}  
\caption{Use EMA to smooth border and noise samples' hardness. The figure shows the hardness curves of border and noise samples, where the x-axis represents training epochs and the y-axis represents hardness values. Compared to using the hardness value from the final epoch directly, our EMA-smoothed mechanism yields more concentrated hardness distributions at the final epoch and enhances the discriminability between noise and border samples.}
\label{fig:intra-cross-0912}
\end{figure*}

\subsection{Comparison with Existing Methods}
\label{subsec:comparison}

We compare our framework with representative methods covering reweight, resample and ensemble paradigms. All methods are model-agnostic and adopt the same base model in Section \ref{sec:model}. Reweight Methods: Focal Loss \cite{lin2017focal} and Asymmetric Loss \cite{ridnik2021asymmetric}; Resample Methods: SMOTE \cite{chawla2002smote} and ENN \cite{wilson1972asymptotic}. Ensemble Methods: EasyEnsemble \cite{liu2008exploratory}, DeepEnsemble \cite{lakshminarayanan2017simple} and the SOTA method SADE \cite{zhang2022self} in \cite{zhang2023deep}; To ensure fair comparisons, we evaluate the above methods with three different sets of hyperparameters and report their best performance. For SADE, we use non-focal mask as augmentation for its self-supervised aggregation.

\begin{table}[h]
    \centering
    \caption{Performance Comparison of Different Methods}
    \label{tab:method_comparison} 
    \resizebox{1.0\linewidth}{!}{
        \begin{tabular}{l c c} 
            \toprule
            Method                                   & DP@R90\% & FP@R90\% \\
            \midrule
            Base Model     & 44.2\%     & 45.7\%     \\
            \midrule
            Focal Loss \cite{lin2017focal}        & 46.5\%     & 45.3\%     \\
            ASL \cite{ridnik2021asymmetric}                                   & 48.2\%     & 46.8\%     \\
            \midrule
            SMOTE \cite{chawla2002smote}             & 45.9\%     & 44.1\%     \\
            ENN \cite{wilson1972asymptotic}                                  & 47.3\%     & 45.5\%     \\
            \midrule
            DeepEnsemble \cite{lakshminarayanan2017simple}                          & 49.8\%     & 48.6\%       \\
            EasyEnsemble \cite{liu2008exploratory}                          & 33.8\%     & 32.1\%   \\
            SADE \cite{zhang2022self}                                  & 44.2\%     & 41.7\%     \\
            \midrule
            \textbf{Ours} & 60.1\%     & 59.7\%     \\
            \bottomrule
        \end{tabular}
    }
    \\[6pt] 
\end{table}

\textbf{Experimental Results:}
As shown in Table \ref{tab:method_comparison}, the reweight methods provide moderate improvements as their general-purpose solutions but show limited effectiveness in handling the extreme imbalance and asymmetric noise in AEB data. The resample methods like SMOTE and ENN, while successful in traditional domains, suffer from domain-specific limitations in AEB scenarios where samples have complex temporal-spatial patterns of driving events. Among ensemble methods, EasyEnsemble's aggressive balancing strategy discards substantial amounts of majority class data, leading to degraded performance. SADE, designed specifically for image classification, lacks the generality required for multi-modal AEB data. In contrast, DeepEnsemble demonstrates more stable performance through its straightforward ensemble diversity, making it the strongest baseline. 

Finally, our method achieves the best performance, which improves 10.3\% DP@R90\% and 11.1\% FP@R90\% compared to the suboptimal DeepEnsemble, demonstrating substantial improvements over all baseline methods.


\subsection{Ablation Study}
\label{subsec:ablation}

In this section, we conduct ablation experiments to validate the impact of different upsample ratios from generated delayed/false trigger and our noise suppression method.

\begin{table}[h]
    \centering
    \caption{Performance Comparison of Our proposed Methods}
    \label{tab:sampling_perf}
    \resizebox{1.0 \linewidth}{!}{
        \begin{tabular}{l c c} 
            \toprule
            Method                & DP@R90\% & FP@R90\% \\
            \midrule
            Deep Ensemble           & 49.8\%     & 48.6\%     \\
            OS@0.5 Ensemble         & 53.3\%     & 51.7\%     \\
            OS@1 Ensemble           & 56.5\%     & 55.3\%  \\
            OS@2 Ensemble           & 55.2\%     & 54.9\%  \\
            OS@1+NS Ensemble        & 60.1\%     & 59.7\%     \\
            \bottomrule
        \end{tabular}
    }
    \\[6pt] 
    \begin{minipage}{1.0 \linewidth}
        \small \textit{Note:} OS = Oversample delayed/false triggers from augmentated data; NS = Noise suppression for true triggers. OS@0.5 represents the ratio of synthetic minority class data to the real minority class data is 0.5:1. All ensemble methods are based on 3 models.
    \end{minipage}
\end{table}

\textbf{Impact of Augmentation Ratios:}
We generate approximately 11,000 synthetic minority samples using our three augmentation strategies with equal proportions. From this pool, we sample different ratios to investigate optimal augmentation levels.

As shown in Table \ref{tab:sampling_perf}, OS@0.5 (0.5 generated : 1 real) provide 3.5\%/3.1\% improvement, which demonstrates that even conservative augmentation helps. OS@1 improves to 56.5\%/55.3\%, showing that balanced augmentation further enhances performance. OS@2 reaches 55.2\%/54.9\%, slightly lower than OS@1, suggesting that excessive synthetic samples may introduce distribution shift. The optimal 1:1 ratio balances between addressing data scarcity and maintaining authentic minority class patterns.

\textbf{Effectiveness of Noise Suppression:}
Our hardness-based noise detection identifies approximately 0.75\% of majority class samples exceeding the adaptive threshold. Adding this noise suppression to the optimal augmentation (OS@1+NS) achieves 3.6\%/4.4\% additional gain. This validates that removing mislabeled majority samples significantly improves the model's ability to learn minority class patterns.

These results confirm that our minority augmentation for minority and noise suppression for majority work synergistically to maximize delayed/false trigger detection performance, fulfilling AEB annotation's core objective.

\subsection{Hardness Stability Analysis}
\label{subsubsec:hardness_exp}

To further validate our proposed hardness-based noise separation mechanism in Section \ref{subsec:noise_suppression}, we conduct both qualitative and quantitative analyses. We select 50 border samples based on model performance on validation set and manually create 50 noisy samples by deliberately mislabeling delayed/false triggers as true triggers.

\begin{table}[h]
    \centering
    \caption{Overlap Coefficient Between Border and Noise Samples}
    \label{tab:hardness_analysis}
    \resizebox{1.0\linewidth}{!}{%
        \begin{tabular}{l c}
            \toprule
            Method & Overlap Coefficient \\
            \midrule
            Raw Hardness & 0.305 \\
            Mean Hardness & 0.124 \\
            EMA Hardness@0.01 & 0.075 \\
            EMA Hardness@0.1 & 0.130 \\
            EMA Hardness@0.05 & 0.050 \\
            \bottomrule
        \end{tabular}
    }
    \\[6pt] 
    \begin{minipage}{1.0 \linewidth}
        \small \textit{Note:} EMA Hardness@0.05 means that EMA parameter $\alpha$ is 0.05.
    \end{minipage}
\end{table}

\textbf{Qualitative Analysis:} As illustrated in Figure \ref{fig:intra-cross-0912}. The \textit{Raw} curves represent the hardness values computed by the model at last epoch, while the \textit{Smooth} curves show the hardness distributions obtained through Exponential Moving Average (EMA) that incorporates the complete training dynamics. Several key observations emerge: (1) Raw hardness values exhibit a large variance, making it difficult to distinguish between border and noise samples; 
(2) The model may overfit some noise samples and give a very small hardness result in final epoch. (3) After applying intra-model smoothing, the mean hardness values ($\mu$) of border and noise samples become more separable, with reduced variance ($\sigma^2$) indicating improved stability; 

\textbf{Quantitative Analysis:} We measure separability using overlap coefficient—the proportion of samples falling within the overlapping range between border and noise distributions. As shown in Table \ref{tab:hardness_analysis}, overlap coefficient progressively decreases from 0.305 (raw) to 0.124 (mean smoothing) to 0.050 (EMA@0.05), demonstrating that our temporal smoothing effectively distinguishes mislabeled samples.

\section{Deployment as a Self-Evolving Annotation System}

Due to the practical requirements of AEB system optimization, we developed and deployed a system centered around our proposed annotation model.

\textbf{System Architecture and Configuration:} Our system utilizes a cloud-based GPU inference platform with 4× NVIDIA A10 cards running PyTorch Distributed Data Parallel (DDP) framework for model inference. The system supports high-throughput processing with a concurrency of 100 connections, achieving 2000 QPS (Queries Per Second), and manages million-scale data storage capabilities. During inference, we set batch size to 64 with 128 workers, which consumes around 8000×4 MB of GPU memory. The front-end utilizes Vue3 framework while the backend employs FastAPI-Python architecture for efficient API services. 

\begin{figure}[h]  
\centering
\includegraphics[width=\linewidth]{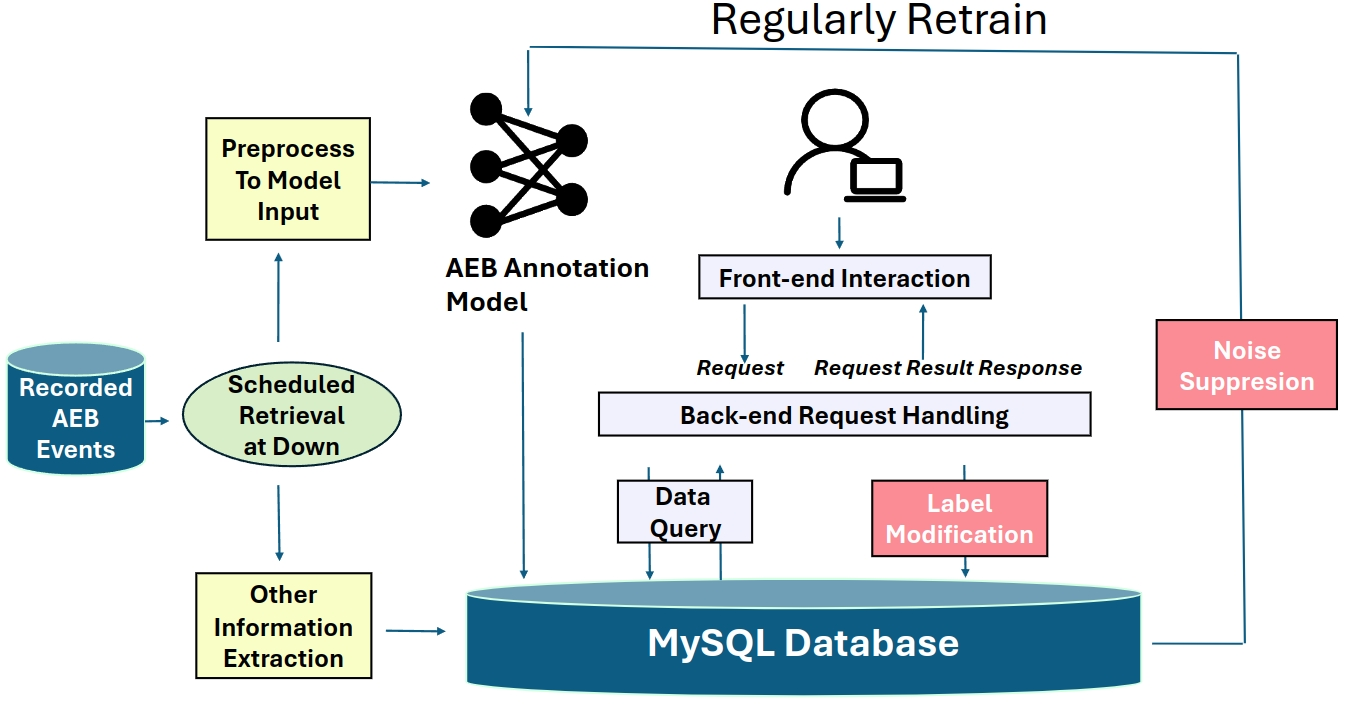}  
\caption{The Architecture of  AEB annotation platform based on our proposed model.}
\label{fig:tower}
\end{figure}

\begin{figure}[h]  
\centering
\includegraphics[width=\linewidth]{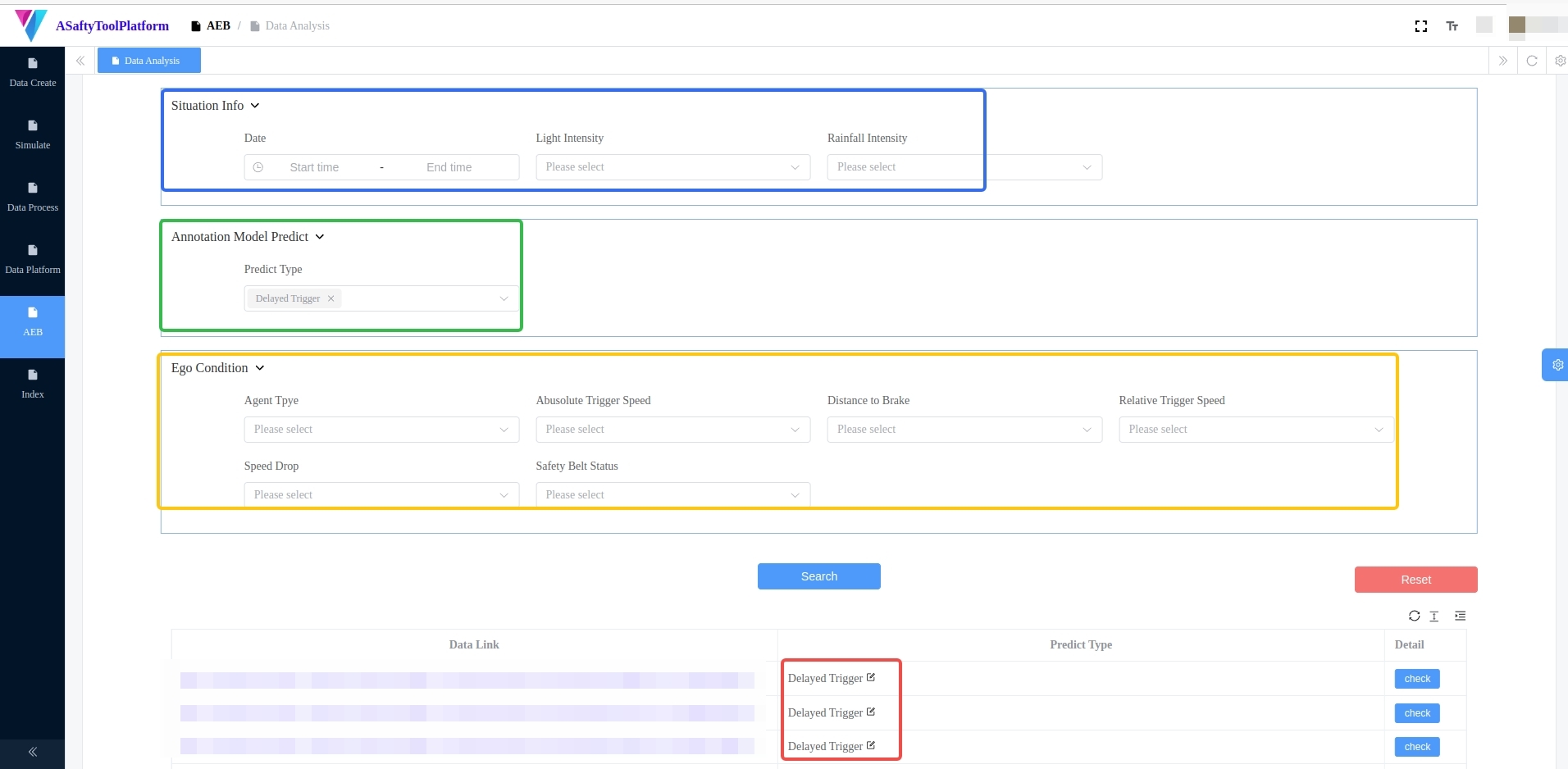}  
\caption{The online Front-end Interface platform. The \textbf{blue panel} enables filtering by time and environmental conditions; the \textbf{green panel} allows analysts to quickly filter samples by model-predicted trigger types (delayed/false/true triggers); the \textbf{yellow panel} provides detailed state filtering including agent type (focal target), trigger speeds, braking distance, and safety belt status—all extracted directly from vehicle data streams. The lower section displays filtered results with direct links to detailed sensor data for each event. Critically, the \textbf{red-boxed labels} support manual correction. }
\label{fig:platform}
\end{figure}

\textbf{Automated Data Pipeline:} The system implements a human-in-the-loop architecture that combines automated efficiency with human expertise. The detailed pipeline is shown in Figure \ref{fig:tower}: 
(1) Scheduled retrieval at dawn of recently recorded AEB events;
(2) Data parsing and preprocessing, which is divided into two parts: first, extracting key information of ego and agents to vectorized inputs required by the model; then, extracting other information such as trigger time, weather, braking distance, etc., and storing them in the database;
(3) Inference by our AEB annotation model to assign labels to each piece of data, and then storing the labeled results. The above three processes can be completed within an hour to handle thousands of AEB trigger samples, ensuring all results are prepared in advance for human review;
(4) Analysts query delayed or false trigger cases through the front-end interface to correct labels and manually analyze the underlying causes. The front-end interface is shown as Figure \ref{fig:platform}.

This closed‑loop system can greatly assist engineers in pinpointing the current shortcomings of AEB, thereby facilitating its optimization and ultimately enhancing the protection of drivers’ and passengers’ safety and property.

\textbf{Self-Evolution Learning:} 
As shown in Figure \ref{fig:tower}, the platform implements a self-evolution mechanism through dual-track data quality enhancement: For minority class, human experts' verification provides high-quality labels for these safety-critical events. For majority class, our hardness-based noise suppression automatically filters mislabeled samples from the massive majority class, maintaining data quality at scale without human intervention.

This dual-track approach creates a virtuous cycle: The accumulated high-quality data enables periodic model retraining, achieving sustained performance improvements.

\textbf{Performance Validation:} 
Following several months of operational deployment, the system demonstrated substantial performance gains: \textbf{80\%} increase in recall quantities for delayed trigger and false trigger scenarios, accompanied by a \textbf{50\%} reduction in manual review overhead. The automated pipeline processes thousands of daily samples with less than 100 samples requiring human verification (95\% automation rate), ultimately accelerating AEB system optimization.

\textbf{Lessons Learned from Practical Model Deployment:}
(1) Model-Reality Gap: Even high-performing models cannot achieve 100\% accurate recall for delayed/false triggers. This inherent limitation motivates our human-in-the-loop design rather than pursuing unrealistic full automation.
(2) Upstream Dependency Drift: AEB annotation relies on upstream perception systems that frequently update. Without retraining, model performance degraded ~1.5\% after 3-4 months due to shifting feature distributions. However, quarterly retraining with accumulated data consistently recovers and improves performance by 2-3\%, turning a maintenance necessity into continuous improvement opportunity.

\section{Conclusion}

In this paper, we present the first automated annotation framework for AEB trigger events, which identifies and solves two fundamental challenges that have hindered delayed/false AEB events annotation: extreme class imbalance and asymmetric label noise.

Our dual-strategy approach combines domain-specific augmentation with probe-guided noise suppression. The augmentation strategies leverage AEB physics to synthesize realistic minority samples, while the hardness-based noise detection adaptively identifies and removes mislabeled majority samples. Experimental results validate that our methods achieve the best 60.1\% and 59.7\% precision at 90\% recall for delayed and false triggers respectively.

We have successfully deployed our model as a production annotation system, demonstrating a 95\% automation rate while maintaining high accuracy for safety-critical events. Importantly, the platform creates a virtuous cycle where accumulated high-quality annotations enable periodic model improvements, transforming a static tool into a continuously improving system.

While focused on AEB, our annotation system's principles extend naturally to other active safety functions. The data challenges we encountered and our dual-strategy solutions are universal across automotive safety such as Emergency Lane Keeping (ELK) and Rear Automatic Emergency Braking (RAEB).
Our work provides not just a tool for AEB but a transferable blueprint for building scalable data foundations essential for all safety-critical automotive systems.

\addtolength{\textheight}{-12cm}   








\bibliographystyle{IEEEtran}
\bibliography{references}

\end{document}